\documentclass[submission,copyright,creativecommons]{eptcs}
\providecommand{\event}{ICLP DC 2023} % Name of the event you are submitting to
\usepackage{graphicx}
\usepackage{iftex}
\usepackage{amsmath}
\usepackage{comment}
\ifpdf
  \usepackage{underscore}         % Only needed if you use pdflatex.
  \usepackage[T1]{fontenc}        % Recommended with pdflatex
\else
  \usepackage{breakurl}           % Not needed if you use pdflatex only.
\fi
\newtheorem{example}{Example}

\title{Penalization Framework For Autonomous Agents Using Answer Set Programming}
\author{Vineel S. K. Tummala
\institute{Computer Science and Software Engineering\\
Miami University\\
Oxford, USA}
\email{tummalvs@miamioh.edu}
}
\def\titlerunning{Penalization Framework For Autonomous Agents Using Answer Set Programming}
\def\authorrunning{Vineel S. K. Tummala}
\begin{document}
\maketitle
\begin{abstract}
This paper presents a framework for enforcing penalties on intelligent agents that do not comply with authorization or obligation policies in a changing environment. A framework is proposed to represent and reason about penalties in plans, and an algorithm is proposed to penalize an agent's actions based on their level of compliance with respect to authorization and obligation policies. Being aware of penalties an agent can choose a plan with a minimal total penalty, unless there is an emergency goal like saving a human's life. The paper concludes that this framework can reprimand insubordinate agents.
\end{abstract}

\section{Introduction and Motivation}

A framework will be presented that can enforce penalties on intelligent agents in a changing environment if autonomous agents don't obey authorization or obligation policies. 
%A discussion is initiated about what intelligent agents are and an authorization policy is described. 
Policies for an agent $\textit{A}$ acting in a changing domain/environment $\textit{T}$ are encoded as a set of conditions 
 that describes whether an agent's actions are permitted or not, and whether the agent is obligated to perform an action or not perform it. In some cases, an autonomous agent can choose to perform an unauthorized action.  In this case, the agent is forced to pay a penalty, where it can be reprimanded for its actions or lose its job for insubordination~\cite{blount2013architecture}. To make this possible, an architecture will be created to represent and reason about penalties in plans.
 
 As autonomous agents continue to develop, it is necessary that proper policy representation and comprehension exist. It is essential to monitor the issues within the environment to ensure that agents act appropriately. Autonomous agents are intended to perform certain actions without any instructions or interference from a controller (pilot or driver) of the agent. The activities of an autonomous agent are often modeled after human behavior so that the agent can perform tasks that humans would. In order to do this, human behavior needs to be accurately represented. This can be successfully achieved by a logical approach, using declarative programming languages. Declarative programming languages are able to define human behavior correctly and in an understandable way. One such declarative programming language is Answer Set Programming~\cite{DBLP:journals/cacm/BrewkaET11}, 

 %these two concepts intersect. Interference between Authorization and Obligations can create loopholes, and exceptions, or even can create a policy that is not compliant.
  
\section{Background and overview of the existing literature}
In this section, background work related to the penalization framework for autonomous agents will be described. I assume that the reader is familiar with Answer Set Programming (ASP) \cite{DBLP:conf/iclp/GelfondL90,DBLP:conf/iclp/GelfondL88}, a declarative programming language with roots in Logic Programming and Nonmonotonic reasoning~\cite{DBLP:conf/rweb/EiterIK09}, and the ASP solver {\sc Clingo}\footnote{\url{https://potassco.org/Clingo/}}. Full details about \textsc{Clingo} can be found in the papers by Calimeri et al. ~\cite{calimeri2020asp} and Gebser et al.~\cite{GebserKKOSS11}.
%\footnote{https://citeseerx.ist.psu.edu/document?repid=rep1\&type=pdf\&doi=eb2b4e1ccb8449353e9adb568d1017fb1be1c1f9}
I also assume familiarity of the reader with autonomous intelligent agents.
 
\subsection{Transition Systems}
In order for intelligent agents to be functional and autonomous, they require a proper description of the changing environment in which they act. Transition systems are a tool that is used to describe these domains. %Dynamic environments are the environments that will be changed based on the actions that take place in said environment~\cite{blount2013architecture}. 
Transition systems are fundamentally directed graphs that represent a changing domain, in which the nodes represent the state of the domain, and actions that might occur in the domain are represented by arcs~\cite{blount2013architecture}. As a result, the domain is altered from one state to another. %Theoretically 
Transition diagrams can become huge if many fluents are being tracked, thus they are not used in practice. They can be represented in a more concise way by an action language~\footnote{\label{foot}See \url{https://www.diva-portal.org/smash/get/diva2:1716299/FULLTEXT01.pdf}} which is a high-level declarative language.\vspace{1.5mm}

\subsection{Policies}

The proposed research will be focusing on agents which are aware of policies. Policies are sets of rules to follow in a given domain or environment and are also described using declarative programming languages. Given a policy, P, with respect to a given environment T, all actions performed in the said environment can be judged based on whether or not they follow the rules defined by the policy. In the case of an intelligent agent, it is important for it to be aware of all the policies in advance in order to comply with all the necessities when coming up with plans for action. %Because of this, policies are defined through declarative programming languages in order to give agents a better knowledge of the norms in their domain/environment. 
The following section will define compliance to policy, look into both authorization and obligation policies.

\subsubsection{Authorization} Authorization policies are policies in which an agent is either permitted (authorized) or not permitted (not authorized) to perform an action. %~\cite{DBLP:conf/iclp/GelfondL08}. 
Gelfond and Lobo~\cite{DBLP:conf/iclp/GelfondL08} define Authorization Policy Language ($\textit{APL}$) to describe authorization policies in a dynamic system  $\Sigma = \langle A, T \rangle$, with an agent \textit{A} and a transition diagram/environment \textit{T}. This language is implemented through Answer Set Programming (ASP). The signature of \textit{APL} includes a predicate for authorization policies \textit{permitted(e)} where \textit{e} is an elementary action of $\textit{A}$.~\cite{DBLP:conf/iclp/GelfondL08}.

\subsubsection{Obligation} 
An obligation policy is one in which an agent is obligated to carry out an action or not to carry it out. \textit{AOPL} is a policy language to describe systems with an agent \textit{A} and a transition diagram \textit{T} over signature $\Sigma$, similar to \textit{APL}~\cite{DBLP:conf/iclp/GelfondL08}. Known as \textit{Authorization and Obligation Policy Language}, Gelfond and Lobo~\cite{DBLP:conf/iclp/GelfondL08} designed this language, to extend \textit{APL} with obligation policies by adding a new symbol \textit{obl(h)} where \textit{h} is an elementary action of $\textit{A}$ or negation of any such action~\cite{DBLP:conf/iclp/GelfondL08}. This language is implemented through ASP.

\subsection{Compliance}
Compliance denotes whether an action or set of actions should be taken or not in accordance with a given policy $\textit{P}$. A set of actions is defined as strongly compliant, weakly compliant, or non-compliant. Strongly compliant means that no rules were broken and it allows/supports their execution. If there is no sufficient information then it is called weakly compliant, and if the actions have broken one or more policies then it is called as non-compliant. 
\subsubsection{Authorization Compliance}

%TODO: need to define event, compound action, and \textit{P($\sigma$)}

Compliance in relation to authorization policies, as defined by \textit{APL($\Sigma$)}, can be divided into strongly compliant, weakly compliant, or not compliant~\cite{DBLP:conf/iclp/GelfondL08}. 

\subsubsection{Obligation Compliance}
For obligation policies defined through \textit{AOPL}($\Sigma$), events are classified into \textit{compliant} or \textit{non-compliant}~\cite{DBLP:conf/iclp/GelfondL08}. Unlike authorization policies, there are no weakly compliant  events with respect to obligations. 

\subsection{Planning}
A plan is a sequence of actions taken by an agent in order to achieve a given goal~\cite{gelfond2014knowledge}. %In relation to intelligent agents Hayes et al~\cite{hayes1993plans} claim that plans are descriptions given to an agent to behave as intended. The purpose of following a plan is to determine the course of behavior, while ``plan following'' is to choose the actions that are compatible with a said plan~\cite{gelfond2014knowledge}.
\textit{Answer Set Planning}~\cite{DBLP:conf/lpnmr/Lifschitz99} refers to the use of Answer Set Programming to solve different types of planning problems by adding a planning module to an ASP description of a changing environment. Answer sets of the resulting program correspond to possible plans.

\section{Goal of the research}
The goal of the proposed research is to create a penalization framework for autonomous agents using ASP where autonomous agents are penalized if they do any job/task that they are not intended to do according to the policy.
\subsection{Constructing the Framework}
In this section, we present our ideas for constructing a penalization framework for autonomous agents. 
 % ideas for constructing a penalization framework for autonomous agents are presented in this section by us.
    % the tasks are related to research questions at the top 
\subsubsection{Encoding the penalties}
    In order for these penalties to be imposed on the agents we will create a statement to penalize the agents. we will choose a specific scale to impose penalty points for agents; as of now the scale is 1-3, in the actual framework we will explore different possibilities. 

We keep track of the values as low, medium, and high to help the agent calculate the value of the penalty while choosing the plans.
% The values are kept track of as low, medium, and high to assist the agent in calculating the penalty value while selecting plans.
    %paper 

\subsubsection{Calculating the total penalty}
We calculate the total penalty, in the end, using the aggregate sum in \textsc{Clingo}\footnote{\url{https://potassco.org/Clingo/run/}} (solver for ASP). Aggregates are functions that combine the values of a set of terms to produce a single value. Using this function we add up all the penalties given to an agent in an environment and we can evaluate an agent based on these penalties.
% The total penalty is calculated in the end, using the aggregate sum in Clingo.

\begin{comment}
\begin{example}
Assume we have a predicate $value(X, Y)$ representing values in a list, where $X$ is the index in the list and $Y$ is the value. Now, we try to add all the values in the list.

% Input facts representing 10 values

$$
\begin{array}{l}
value(1, 3).\\
value(2, 5).\\
value(3, 7).\\
value(4, 2).\\
value(5, 1).\\
value(6, 4).\\
value(7, 9).\\
value(8, 8).\\
value(9, 6).\\
value(10, 0).
\end{array}
$$

Compute the sum of the values using the aggregate and sum operators
$$sum\_values(S) :- \#sum \{ V : value(\_, V) \} = S.$$

 Show the sum of the values
$$\#show \ sum\_values/1.$$
\end{example}
\end{comment}

    % aggregates and sum'

\section{Current status of the research}
Collected all background and related work, and created two domains to test the penalization framework.
\subsection{Collecting policies and scenarios}
When we speak about penalizing autonomous agents, we need to create domains and scenarios where we evaluate autonomous agents for their actions and penalize them. Also, to achieve this we need to specify authorization and obligation policies for the domain beforehand. Conflicts may arise when Authorization and obligation policies interact or come across each other. Therefore we need to specify authorization and obligation policies precisely. First definitions and rules of both the authorization and obligation policies developed by Gelfond and Lobo~\cite{DBLP:conf/iclp/GelfondL08} must be understood. After fully understanding the rules that exist, different scenarios can be tested. Scenarios will be tested to see what an agent selects if it has two non-compliant plans, and what plan it chooses.
% When penalizing autonomous agents is discussed, domains and scenarios need to be created where the actions of autonomous agents are evaluated and penalized. Moreover, authorization and obligation policies for the domain need to be specified beforehand in order to achieve this.
Some of the scenarios are:
\subsubsection{Domain 1: Drone Delivery}
\textbf{Goal:}
Its goal is to deliver a package to a customer's doorstep, located in a suburban area with many houses.

\smallskip
%\medskip
\noindent
\textbf{Policy Rules:} Consider an autonomous agent in this scenario as a delivery drone with the following rules to ensure safe and efficient deliveries:
    \begin{itemize}
        \item Always fly at a safe height to avoid obstacles and people on the ground.
        \item only land on designated landing pads or open spaces to avoid collisions and ensure safe deliveries.
    \end{itemize}

% similarly these scenarios will be used in the evaluation of the framework 

%we are using domain 1 for guiding the construction of framework and other two will be for evaluating 
% how to encode penalties by extending aopl
% calculated penalty how is it actually useful in planning 
% how does the strength of the goal effects the decision making for an agent 
\noindent

\subsubsection{Domain 2: Self-Driving Car} The autonomous agent in this scenario is a self-driving car, programmed with a set of rules to ensure safe and efficient transportation.

\vspace{2.5mm}
\noindent
\textbf{Goal:} Its goal is to transport a passenger from their current location to a specified destination, which involves driving through a busy city street with heavy traffic.

\smallskip
%\medskip
\noindent
\textbf{Policy Rules:} Consider an autonomous agent in this scenario as a self-driving car with the following rules to ensure safe and efficient deliveries:
    \begin{itemize}
        \item Always obey traffic laws and signals, including speed limits, stop signs, and traffic lights.
        \item Prioritize the safety of passengers and other drivers on the road.
    \end{itemize}
    
\noindent

\section{Open issues and expected achievements}
\textbf{Open Issues}
\begin{itemize}
    \item We can't create a penalty for each and every possible action, so should we set a default value if the action is not encoded?
    \item Setting a maximum penalty limit after which the agent would be automatically disabled from its operations for not following the compliance. This would ensure the agents that they do not perform any operations that are not compliant. 
\end{itemize}

The expected achievement is to create a framework where autonomous agent chooses plans with fewer penalties and also according to the strength of the goal.
\bibliographystyle{eptcs}
\bibliography{bibliography}

\begin{thebibliography}{10}
\providecommand{\bibitemdeclare}[2]{}
\providecommand{\surnamestart}{}
\providecommand{\surnameend}{}
\providecommand{\urlprefix}{Available at }
\providecommand{\url}[1]{\texttt{#1}}
\providecommand{\href}[2]{\texttt{#2}}
\providecommand{\urlalt}[2]{\href{#1}{#2}}
\providecommand{\doi}[1]{doi:\urlalt{https://doi.org/#1}{#1}}
\providecommand{\eprint}[1]{arXiv:\urlalt{https://arxiv.org/abs/#1}{#1}}
\providecommand{\bibinfo}[2]{#2}

\bibitemdeclare{phdthesis}{blount2013architecture}
\bibitem{blount2013architecture}
\bibinfo{author}{Justin \surnamestart Blount\surnameend}
  (\bibinfo{year}{2013}): \emph{\bibinfo{title}{An architecture for intentional
  agents}}.
\newblock Ph.D. thesis, \doi{10.4204/EPTCS.345.23}.

\bibitemdeclare{article}{DBLP:journals/cacm/BrewkaET11}
\bibitem{DBLP:journals/cacm/BrewkaET11}
\bibinfo{author}{Gerhard \surnamestart Brewka\surnameend},
  \bibinfo{author}{Thomas \surnamestart Eiter\surnameend} \&
  \bibinfo{author}{Miroslaw \surnamestart Truszczynski\surnameend}
  (\bibinfo{year}{2011}): \emph{\bibinfo{title}{Answer set programming at a
  glance}}.
\newblock {\slshape \bibinfo{journal}{Commun. {ACM}}}
  \bibinfo{volume}{54}(\bibinfo{number}{12}), pp. \bibinfo{pages}{92--103},
  \doi{10.1145/2043174.2043195}.

\bibitemdeclare{article}{calimeri2020asp}
\bibitem{calimeri2020asp}
\bibinfo{author}{Francesco \surnamestart Calimeri\surnameend},
  \bibinfo{author}{Wolfgang \surnamestart Faber\surnameend},
  \bibinfo{author}{Martin \surnamestart Gebser\surnameend},
  \bibinfo{author}{Giovambattista \surnamestart Ianni\surnameend},
  \bibinfo{author}{Roland \surnamestart Kaminski\surnameend},
  \bibinfo{author}{Thomas \surnamestart Krennwallner\surnameend},
  \bibinfo{author}{Nicola \surnamestart Leone\surnameend},
  \bibinfo{author}{Marco \surnamestart Maratea\surnameend},
  \bibinfo{author}{Francesco \surnamestart Ricca\surnameend} \&
  \bibinfo{author}{Torsten \surnamestart Schaub\surnameend}
  (\bibinfo{year}{2020}): \emph{\bibinfo{title}{ASP-Core-2 input language
  format}}.
\newblock {\slshape \bibinfo{journal}{Theory and Practice of Logic
  Programming}} \bibinfo{volume}{20}(\bibinfo{number}{2}), pp.
  \bibinfo{pages}{294--309}, \doi{10.1017/S1471068419000450}.

\bibitemdeclare{inproceedings}{DBLP:conf/rweb/EiterIK09}
\bibitem{DBLP:conf/rweb/EiterIK09}
\bibinfo{author}{Thomas \surnamestart Eiter\surnameend},
  \bibinfo{author}{Giovambattista \surnamestart Ianni\surnameend} \&
  \bibinfo{author}{Thomas \surnamestart Krennwallner\surnameend}
  (\bibinfo{year}{2009}): \emph{\bibinfo{title}{Answer Set Programming: {A}
  Primer}}.
\newblock In \bibinfo{editor}{Sergio \surnamestart Tessaris\surnameend},
  \bibinfo{editor}{Enrico \surnamestart Franconi\surnameend},
  \bibinfo{editor}{Thomas \surnamestart Eiter\surnameend},
  \bibinfo{editor}{Claudio \surnamestart Gutierrez\surnameend},
  \bibinfo{editor}{Siegfried \surnamestart Handschuh\surnameend},
  \bibinfo{editor}{Marie{-}Christine \surnamestart Rousset\surnameend} \&
  \bibinfo{editor}{Renate~A. \surnamestart Schmidt\surnameend}, editors:
  {\slshape \bibinfo{booktitle}{Reasoning Web. Semantic Technologies for
  Information Systems, 5th International Summer School 2009, Brixen-Bressanone,
  Italy, August 30 - September 4, 2009, Tutorial Lectures}}, {\slshape
  \bibinfo{series}{Lecture Notes in Computer Science}} \bibinfo{volume}{5689},
  \bibinfo{publisher}{Springer}, pp. \bibinfo{pages}{40--110},
  \doi{10.1007/978-3-642-03754-2\_2}.

\bibitemdeclare{article}{GebserKKOSS11}
\bibitem{GebserKKOSS11}
\bibinfo{author}{Martin \surnamestart Gebser\surnameend},
  \bibinfo{author}{Benjamin \surnamestart Kaufmann\surnameend},
  \bibinfo{author}{Roland \surnamestart Kaminski\surnameend},
  \bibinfo{author}{Max \surnamestart Ostrowski\surnameend},
  \bibinfo{author}{Torsten \surnamestart Schaub\surnameend} \&
  \bibinfo{author}{Marius \surnamestart Schneider\surnameend}
  (\bibinfo{year}{2011}): \emph{\bibinfo{title}{Potassco: The Potsdam Answer
  Set Solving Collection}}.
\newblock {\slshape \bibinfo{journal}{{AI} Commun.}}
  \bibinfo{volume}{24}(\bibinfo{number}{2}), pp. \bibinfo{pages}{107--124},
  \doi{10.3233/AIC-2011-0491}.

\bibitemdeclare{book}{gelfond2014knowledge}
\bibitem{gelfond2014knowledge}
\bibinfo{author}{Michael \surnamestart Gelfond\surnameend} \&
  \bibinfo{author}{Yulia \surnamestart Kahl\surnameend} (\bibinfo{year}{2014}):
  \emph{\bibinfo{title}{Knowledge representation, reasoning, and the design of
  intelligent agents: The answer-set programming approach}}.
\newblock \bibinfo{publisher}{Cambridge University Press},
  \doi{10.1017/CBO9781139342124}.

\bibitemdeclare{inproceedings}{DBLP:conf/iclp/GelfondL88}
\bibitem{DBLP:conf/iclp/GelfondL88}
\bibinfo{author}{Michael \surnamestart Gelfond\surnameend} \&
  \bibinfo{author}{Vladimir \surnamestart Lifschitz\surnameend}
  (\bibinfo{year}{1988}): \emph{\bibinfo{title}{The Stable Model Semantics for
  Logic Programming}}.
\newblock In \bibinfo{editor}{Robert~A. \surnamestart Kowalski\surnameend} \&
  \bibinfo{editor}{Kenneth~A. \surnamestart Bowen\surnameend}, editors:
  {\slshape \bibinfo{booktitle}{Logic Programming, Proceedings of the Fifth
  International Conference and Symposium, Seattle, Washington, USA, August
  15-19, 1988 {(2} Volumes)}}, \bibinfo{publisher}{{MIT} Press}, pp.
  \bibinfo{pages}{1070--1080}, \doi{10.2307/2275201}.

\bibitemdeclare{inproceedings}{DBLP:conf/iclp/GelfondL90}
\bibitem{DBLP:conf/iclp/GelfondL90}
\bibinfo{author}{Michael \surnamestart Gelfond\surnameend} \&
  \bibinfo{author}{Vladimir \surnamestart Lifschitz\surnameend}
  (\bibinfo{year}{1990}): \emph{\bibinfo{title}{Logic Programs with Classical
  Negation}}.
\newblock In \bibinfo{editor}{David H.~D. \surnamestart Warren\surnameend} \&
  \bibinfo{editor}{P{\'{e}}ter \surnamestart Szeredi\surnameend}, editors:
  {\slshape \bibinfo{booktitle}{Logic Programming, Proceedings of the Seventh
  International Conference, Jerusalem, Israel, June 18-20, 1990}},
  \bibinfo{publisher}{{MIT} Press}, pp. \bibinfo{pages}{579--597},
  \doi{10.1007/BF03037169}.

\bibitemdeclare{inproceedings}{DBLP:conf/iclp/GelfondL08}
\bibitem{DBLP:conf/iclp/GelfondL08}
\bibinfo{author}{Michael \surnamestart Gelfond\surnameend} \&
  \bibinfo{author}{Jorge \surnamestart Lobo\surnameend} (\bibinfo{year}{2008}):
  \emph{\bibinfo{title}{Authorization and Obligation Policies in Dynamic
  Systems}}.
\newblock In \bibinfo{editor}{Maria~Garcia \surnamestart de~la
  Banda\surnameend} \& \bibinfo{editor}{Enrico \surnamestart
  Pontelli\surnameend}, editors: {\slshape \bibinfo{booktitle}{Logic
  Programming, 24th International Conference, {ICLP} 2008, Udine, Italy,
  December 9-13 2008, Proceedings}}, {\slshape \bibinfo{series}{Lecture Notes
  in Computer Science}} \bibinfo{volume}{5366}, \bibinfo{publisher}{Springer},
  pp. \bibinfo{pages}{22--36}, \doi{10.1007/978-3-540-89982-2\_7}.

\bibitemdeclare{inproceedings}{DBLP:conf/lpnmr/Lifschitz99}
\bibitem{DBLP:conf/lpnmr/Lifschitz99}
\bibinfo{author}{Vladimir \surnamestart Lifschitz\surnameend}
  (\bibinfo{year}{1999}): \emph{\bibinfo{title}{Answer Set Planning
  (Abstract)}}.
\newblock In \bibinfo{editor}{Michael \surnamestart Gelfond\surnameend},
  \bibinfo{editor}{Nicola \surnamestart Leone\surnameend} \&
  \bibinfo{editor}{Gerald \surnamestart Pfeifer\surnameend}, editors: {\slshape
  \bibinfo{booktitle}{Logic Programming and Nonmonotonic Reasoning, 5th
  International Conference, LPNMR'99, El Paso, Texas, USA, December 2-4, 1999,
  Proceedings}}, {\slshape \bibinfo{series}{Lecture Notes in Computer Science}}
  \bibinfo{volume}{1730}, \bibinfo{publisher}{Springer}, pp.
  \bibinfo{pages}{373--374}, \doi{10.1007/3-540-46767-X\_28}.

\end{thebibliography}
\end{document}